\documentclass[5p,times]{elsarticle}

\usepackage{lineno,hyperref}
\modulolinenumbers[5]

\usepackage{tikz}
\newcommand{\Mypm}{\mathbin{\tikz [x=1.4ex,y=1.4ex,line width=.1ex] \draw (0.0,0) -- (1.0,0) (0.5,0.08) -- (0.5,0.92) (0.0,0.5) -- (1.0,0.5);}}%

\usepackage{graphicx}
\usepackage{amsmath}
\usepackage{amssymb}

\usepackage{hyperref} % For hyperlinks in the PDF

\usepackage[normalem]{ulem}

\usepackage{arydshln}

\usepackage{natbib}

\definecolor{mygreen}{RGB}{0, 190, 0}

\journal{Journal of \LaTeX\ Templates}

\biboptions{authoryear}

\hypersetup{draft}
\begin{document}

\begin{frontmatter}

\title{3D Regression Neural Network for the Quantification of Enlarged Perivascular Spaces in Brain MRI}

\author[bigr]{Florian~Dubost
\corref{mycorrespondingauthor}}
\cortext[mycorrespondingauthor]{Corresponding author}
\ead{floriandubost1@gmail.com}

\author[epi]{Hieab~Adams}
\author[bigr]{Gerda~Bortsova}
\author[neuroepi]{M.~Arfan~Ikram}
\author[bigr,tudelft]{Wiro~Niessen}
\author[epi]{Meike~Vernooij}
\author[bigr,dtu]{Marleen~de~Bruijne}
\ead{marleen.debruijne@erasmusmc.nl}

\address[bigr]{Biomedical Imaging Group Rotterdam, Departments of Radiology and Medical Informatics, Erasmus MC - University Medical Center Rotterdam, The Netherland}
\address[epi]{Departments of Radiology and Epidemiology, Erasmus MC - University Medical Center Rotterdam, The Netherlands}
\address[neuroepi]{Departments of Radiology, Epidemiology and Neurology. Erasmus MC - University Medical Center Rotterdam, The Netherlands}
\address[tudelft]{Department of Imaging Physics, Faculty of Applied Science, TU Delft, Delft, The Netherlands}
\address[dtu]{Image Group, Department of Computer Science, University of Copenhagen, Copenhagen, Denmark}

\begin{abstract}
Enlarged perivascular spaces (EPVS) in the brain are an emerging imaging marker for cerebral small vessel disease, and have been shown to be related to increased risk of various neurological diseases, including stroke and dementia. Automated quantification of EPVS would greatly help to advance research into its etiology and its potential as a risk indicator of disease. We propose a convolutional network regression method to quantify the extent of EPVS in the basal ganglia from 3D brain MRI.  We first segment the basal ganglia and subsequently apply a 3D convolutional regression network designed for small object detection within this region of interest. The network takes an image as input, and outputs a quantification score of EPVS. The network has significantly more convolution operations than pooling ones and no final activation, allowing it to span the space of real numbers. We validated our approach using a dataset of 2000 brain MRI scans scored visually. Experiments with varying sizes of training and test sets showed that a good performance can be achieved with a training set of only 200 scans. With a training set of 1000 scans, the intraclass correlation coefficient (ICC) between our scoring method and the expert's visual score was 0.74. Our method outperforms by a large margin - more than 0.10 - four more conventional automated approaches based on intensities, scale-invariant feature transform, and random forest. We show that the network learns the structures of interest and investigate the influence of hyper-parameters on the performance. We also evaluate the reproducibility of our network using a set of 60 subjects scanned twice (scan-rescan reproducibility). On this set our network achieves an ICC of 0.93, while the intrarater agreement reaches 0.80. Furthermore, the automated EPVS scoring correlates similarly to age as visual scoring.
\end{abstract}


\end{frontmatter}

\section{Introduction}

\begin{figure*}[!t]
\centering
\includegraphics[height=4.5cm]{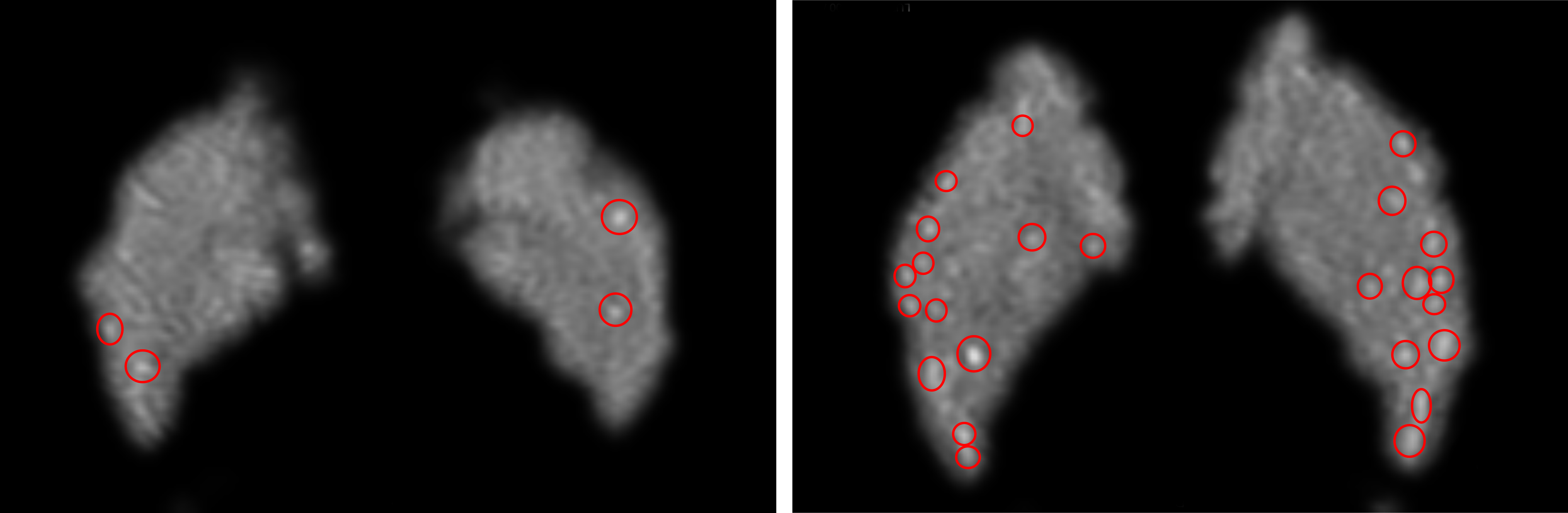}
\caption{\textbf{Examples of enlarged perivascular spaces in the basal ganglia.} EPVS are circled in red. The EPVS have been counted in this slice (Sec. \ref{sec:data}). Note that to correctly identify EPVS, clinicians need to scroll through slices to check the 3D structure of the candidate lesions.}
\label{fig:EPVS}
\end{figure*}

This paper addresses the problem of automated quantification of enlarged perivascular spaces from MR images. The perivascular space - also called Virchow-Robin space - is the space between a vein or an artery and pia mater, the envelope covering the brain. These spaces are known to have a tendency to dilate for reasons not yet clearly understood \citep{Adams2015}. Enlarged - or dilated - perivascular spaces (EPVS) can be identified as hyperintensities on T2-weighted MRI. In Fig. \ref{fig:EPVS}, we show examples of EPVS in T2-weighted scans. Several studies have investigated the presence of EPVS as an emerging biomarker for various brain diseases such as dementia \citep{MILLS2007}, stroke \citep{Selvarajah2009}, multiple sclerosis \citep{Achiron2002} and Parkinson \citep{Zijlmans2004}. In this paper we focus on EPVS located in the basal ganglia. There, the structure of EVPS may for instance relate to the presence or absence of beta-amyloid, a protein that has been implicated in Alzheimer's disease \citep{Pollock1997}. Previous work on automated EPVS quantification focused on the basal ganglia as well \citep{Gonzalez2016,Gonzalez2017}, and clinical studies generally rate the EPVS presence especially in the basal ganglia and centrum semiovale \citep{Wardlaw2013}.

Manual annotation of EPVS is a challenging and very time consuming task: EPVS are thin and small structures - often at the resolution limit of 1.5T and 3T MRI scanners - with much variation in their size and shape. Raters need to zoom and scroll through slices to differentiate EPVS from similarly appearing brain lesions such as lacunar infarcts or small white matter lesions. Additionally, many EPVS can be present within a single scan. In our dataset, for instance, there were up to 35 EPVS within a single slice of the basal ganglia. Current clinical studies rely on visual scoring systems, in which expert human raters count the number of EPVS within a given subcortical structure or region of interest (ROI) \citep{Adams2013, Adams2015} or rate the EPVS on a 5 point scale.

Recently several groups have addressed EPVS quantification using different scenarios and techniques.
\cite{Ramirez2015} developed interactive segmentation methods based on intensity thresholding. 
\cite{Park2016} proposed an automated EPVS segmentation method based on Haar-like features. This approach was exclusively evaluated on 7 Tesla MRI scans and needed a large amount of pixel-wise annotations for training.
\cite{Ballerini2016} used a Frangi filter to enhance EPVS and perform segmentation of individual EPVS. They evaluated their performance using a discrete 5-category EPVS scoring system \citep{Potter2015b}.
In \cite{Gonzalez2016,Gonzalez2017}, in contrast with above approaches, the same authors did not aim to segment individual EPVS. They directly formulated the problem as a binary classification - few or many EPVS - and used bag of words descriptors with support vector machine classification. Our work extends this by proposing, instead of a binary score, a continuous score, translating the presence EPVS.
Recently we published a weakly supervised method using neural networks to detect EPVS in the basal ganglia \citep{Dubost2017}. Our former work targeted a detection problem, and was evaluated with manually annotated EPVS, while in this work we introduce automated EPVS scores without considering the location information, and focus on the evaluation of these scores.  

Our proposed method relies on a 3D regression convolutional neural network (CNN). One of the main advantages of CNN in comparison to other machine learning techniques, is that the features are automatically computed to maximize the final objective function. 3D CNNs have recently received much attention in the medical imaging literature, for instance for segmentation \citep{Chen2017,Bortsova2017,Cicek2016}, landmark detection \citep{Ghesu2016a} or lesion detection \citep{Dou2016}. CNN regression tasks have been less addressed in medical imaging. For instance in \cite{Miao2016}, a set of local 2D CNN regressors are employed for 2D/3D registration. \cite{Xie2016} propose a fully convolutional network to count cells by regressing their 2D density maps generated from dot-annotations.

\textbf{Contributions.} In this paper we propose an automated scoring method to quantify EPVS in the basal ganglia. The method is based on a 3D-CNN for regression problems and uses only visual scores labels for training. This scoring method eases the annotation effort and provides a fine scale quantification. We demonstrate the potential of our method on EPVS in the basal ganglia. We show that our method correlates well with the visual scores of expert human raters and that the correlation of the automated scores with increasing age is similar to that of visual scores. It is the first time that an automated EPVS quantification method is evaluated on such a large dataset (2000 MR scans).

\section{Materials and Methods}
\label{sec:method}

The objective of our method is to automatically reproduce the EPVS visual scores.
Our framework consists of two steps. We first isolate the region of interest (ROI) (Sec\ref{sec:preProcess}) and then apply a regression convolutional neural network (CNN) (Sec\ref{sec:regNet}) to compute the EPVS presence score.

\subsection{Data}
\label{sec:data}
In our experiments we used brain MRI scans from the Rotterdam Scan Study. The Rotterdam Scan Study is an MRI based prospective population study investigating - among others - neurological diseases in the middle aged and elderly \citep{Ikram2015}. The scans used in our experiment were acquired with a GE 1.5 Tesla scanner, between 2005 and 2011. The age of the participants ranges from 60 to 96 years old. 

The scans were visually scored by a single expert rater (H. Adams), who counted - without indicating their location - the number of EPVS in the basal ganglia, in the slice showing the anterior commissure \citep{Adams2015} (see Fig \ref{fig:EPVS} for a few examples). The number of EPVS in this slice correlates with the number of EPVS in the whole volume \citep{Adams2013}. 

\subsubsection{Size of the Datasets}

In total, the visually scored dataset contains 2017 3D MRI scans from 3 different sub-cohorts. From these 2017 scans, 40 scans have also been visually scored by a second trained rater (F. Dubost), and 25 scans have been marked with dot annotations (by H. Adams) at the center of EPVSs to check the focus of the network. Note that only EPVS in the slice showing the anterior commissure have been marked. In addition, we used 46 other scans for which 23 study participants were scanned twice within a short period (19 $\Mypm$ 11 days). The 46 scans of this reproducibility set are not part of the 2017 scans mentioned above and were not visually scored for EPVS. 

\subsubsection{Scans Characteristics}

We used PD-weighted images for our experiments. The scans were acquired according to the following protocol: 12,300 ms repetition time, 17.3 ms echo time, 16.86 KHz bandwidth, 90-180$^{\circ}$ flip angle, 1.6 mm slice thickness, 25 cm$^2$ field of view, $416 \times 256$ matrix size. The images are reconstructed to a  $512 \times 512 \times 192$ matrix. The voxel resolution is $0.49 \times 0.49 \times 0.8 \text{mm}^{3}$. Note that these PD-weighted images have a contrast similar to T2-weighted images, the modality more commonly used to detect EPVS.

\subsubsection{Quality of the Visual Scoring}

Visual EPVS scores have been created according to a standard procedure proposed in the international consortium UNIVRSE \citep{Adams2015}. H. Adams established the UNIVRSE standardized EPVS scoring system and had three years’ experience in identifying EPVS at the moment he annotated the scans for the current study. Intrarater reliability for this scoring has been computed on the Rotterdam Scan Study, and was reported to be excellent in the basal ganglia (Intraclass Correlation Coefficient (ICC) of 0.80 computed on 85 scans) and inter-rater reliability was reported to be good (ICC of 0.62 on 105 scans) \citep{Adams2013}. We plotted a histogram of the EPVS distribution in Fig. \ref{fig:results}.

\begin{figure*}
\centering
\includegraphics[height=3.5cm]{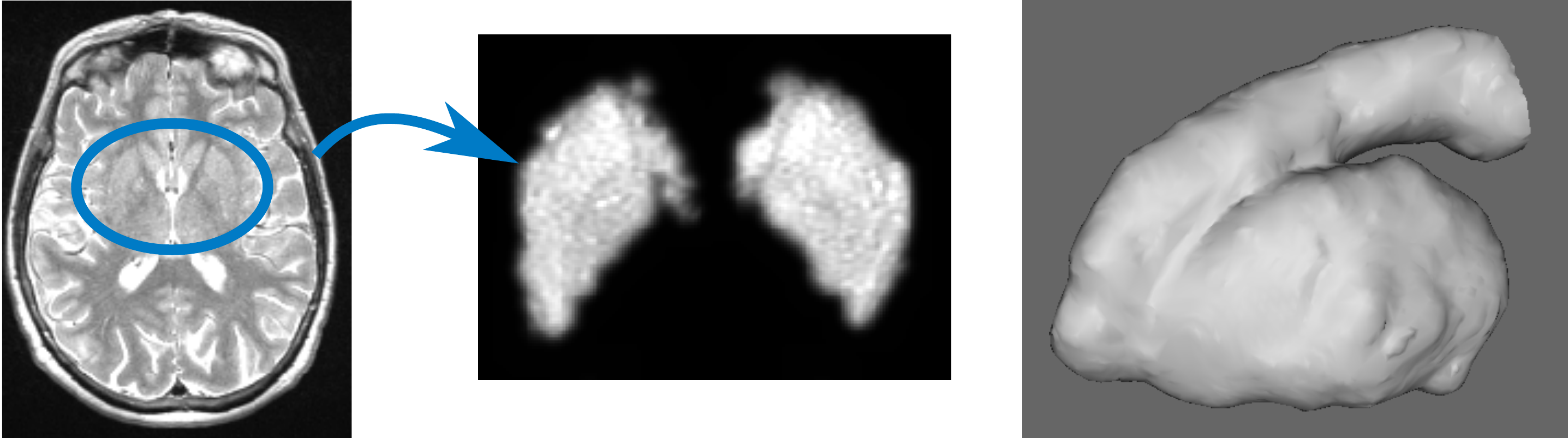}
\caption{\textbf{Preprocessing: computation of a smooth mask of the basal ganglia.} From left to right: full MRI scan in axial view; basal ganglia after computation of the smooth mask; 3D rendering of the basal ganglia.}
\label{fig:preprocessing}
\end{figure*}
\subsection{Preprocessing - Smooth ROI}
\label{sec:preProcess}
We first extract a smooth ROI, which can be seen as a spatial prior and focuses the neural network to a predefined anatomical region. In case of 3D images, computing a ROI also helps avoiding the overload of GPU memory and allows to build deeper networks and to train faster.

A binary mask would arbitrarily impose a hard constraint on the input data and can lead to unwanted border effects. Therefore we propose to compute a smooth mask.

Each scan is first registered to MNI space resulting in the hypermatrix $V \in \mathbb{R}^{H \times W \times D}$. A binary mask of the ROI, $M_{b} \in \{0,1\}^{H \times W \times D}$, is then created using a standard algorithm for subcortical segmentation \citep{Desikan2006}. The mask is then dilated by first applying $4$ consecutive morphological binary dilations with a square connectivity equal to one (6 neighbors in 3D) and subsequently smoothed by convolving the mask with a Gaussian kernel of standard deviation $\sigma$. The dilation ensures that EPVS located at the border of the ROI are not segmented out. The resulting smooth mask $M_{s} \in [0,1]^{H \times W \times D}$ is then multiplied element-wise with the volume $V$, and cropped in all 3 dimensions around its center of mass to get the final preprocessed image $S \in \mathbb{R}^{h \times w \times d}$, with $h \leq H$, $w \leq W$ and $d \leq D$. In the following sections we refer to $S$ as the smooth ROI. See figure \ref{fig:preprocessing} for an illustration of the computation of the smooth ROI.
We rescale $S$ by dividing by the maximum intensity such that $S \in [0,1]^{h \times w \times d}$. This type of intensity standardization has been successfully used in other deep learning frameworks for quantification and detection of brain lesions \cite{Dou2016}.
\subsection{3D Convolutional Regression Network}
\label{sec:regNet}
\begin{figure*}[t]
\centering
\includegraphics[height=2.6cm]{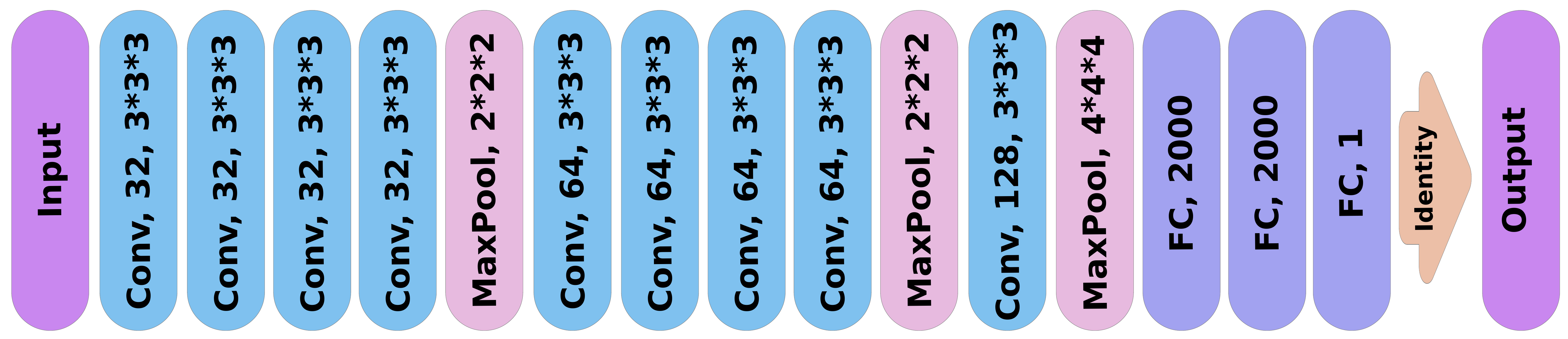}
\caption{\textbf{3D Regression CNN Architecture.} The first two blocks consist of 4 3D convolutions followed by a max-pooling. The last block, before the fully connected layers, only has one convolution followed by a larger max-pooling. After each convolutional layer, we apply a rectified linear unit activation. This architecture is specifically designed to detect small lesions.}
\label{fig:arch}
\end{figure*}
Once the smooth ROI $S$ is computed we use it as input to a convolutional neural network (CNN) which proceeds to the regression task. 

Our CNN architecture is similar to that of VGG \citep{Simonyan14c} but uses 3D convolutional kernels and a single input channel. Additionally, we adapt the architecture for better detection of small structures. We detail our architecture in the following paragraph. Please refer to Fig.\ref{fig:arch} for a visual representation of the network.

The network consists of two blocks of consecutively stacked convolutional layers with small filter size: $3 \times 3 \times 3$, followed by a third block containing a single convolutional layer. We could not expand the network further because of the size of our GPU memory. Note that we do not use any padding and the size of the feature maps is thus reduced after each convolution. Therefore, the input ROI should be sufficiently large to ensure that EPVS located close to its border are not missed. After each convolutional layer we apply a rectified linear unit activation. Between each block of convolutions, a maxpooling layer downsamples the feature maps by $2$ in each dimension (Fig. \ref{fig:arch}). We increase the number of features maps by 2 after each pooling, following the recommendations in \cite{Simonyan14c}. The last pooling layer downsamples its input by $4$. The network ends with two fully connected (FC) layers of $c = 2000$ units and a final FC layer of a single unit. 

As we framed the problem as a regression, the output should span $\mathbb{R}$. The last activation is then only the identity function. The network parameters are optimized using the mean squared error between $y \in \mathbb{N}^{n}$, the EPVS visual scores, and $\hat{y} \in R^{n}$, the output of the network. The EPVS score $\hat{y}$ is therefore optimized to predict the number of EPVS inside the basal ganglia in the slice showing the anterior commissure. However, contrary to an EPVS count, our EPVS scoring can span $\mathbb{R}$ and not only $\mathbb{N}$. The use of a continuous scoring can reflect the uncertainty in identifying a lesion as an EPVS. Besides, the network is regularized only using data-augmentation (Sec. \ref{sec:param}).

Architecture choices can be explained as follows. In the brain there can be different type of lesions appearing similar on a given MRI modality. EPVS are for instance difficult to discriminate from lacunar infarcts on our PD-weighted scans. Therefore complex features should be extracted at high image resolution, before any significant downsampling. For this reason we place the majority of the convolutional layers before and right after the first maxpooling. 
Once these small structures have been detected, there is no need to reach a higher level of abstraction: they only need to be counted. That is our motivation to perform only few pooling operations and finish with a large $4 \times 4 \times 4$ pooling.
The role of the fully connected layers is to estimate the EPVS score based on the EPVS detections provided by the output of the last pooling layer. Ideally the output of the last pooling layer could be a set of low dimensional feature maps highlighting the structures of interest, in our case the EPVS.

%------------------------------------------------

\section{Experiments and Results}
\label{sec:results}

In order to evaluate the performance of the proposed quantification technique, we conduct seven experiments. In the two first experiments we investigate the behavior of the network and check if the network focuses on EPVS. The third series of experiments compares our method with visual scores and with other automated approaches to EPVS quantification. Then we investigate the influence of the number of training samples. In the fifth experiment we analyze the influence of several hyper-parameters on the performance of the network. In the sixth experiment, we assess the reproducibility of our method on short term repeat scans. Finally we show how our EPVS scoring correlates with age.
\subsection{Experimental Settings}
\label{sec:param}
In each experiment the preprocessing is the same (Sec.\ref{sec:preProcess}).
The basal ganglia is segmented with the subcortical segmentation of FreeSurfer \citep{Desikan2006}. All parameters are left as default, except for the skull stripping preflooding height threshold which is set to 10.
Registration to MNI space is computed with the rigid registration implemented in Elastix \citep{Klein2010} and uses default parameters with mutual information as similarity measure. The voxel size stays the same in dimensions x and y (both 0.5mm) but is different in dimension z (0.8mm before registration and 0.5 after). The Gaussian kernel used to smooth the ROI has a standard deviation $\sigma = 2$ pixel units.
The cropped CNN inputs $S$ have a size of $168 \times 128 \times 84$ voxels. We initialize the weights of the CNN by sampling from a Gaussian distribution, use Adadelta \citep{Zeiler2012} for optimization and augment the training data with randomly transformed samples. The transformation parameters are uniformly drawn from an interval of $0.2$ radians for rotation, $2$ pixels for translation and flipping of $x$ and $y$ axes.

The network is trained per sample (mini-batches of a single 3D image).
We implemented our algorithms in Python in Keras and Theano and ran the experiments on a Nvidia GeForce GTX 1070 GPU. This GPU has 8GB of GPU RAM, which prevents us from extending the network.

The average training time is one day. We stop the training after the validation loss converged to a stable value. Once the CNN is trained and given the smooth ROI $S$, the automated EPVS scoring takes $440$ ms on our GPU and 2 min on our CPU.
We evaluate the results using four metrics: the Pearson correlation coefficient, the Spearman correlation coefficient, the intraclass correlation coefficient (ICC) and the mean square error (MSE). We compute these metrics between the visual scores of the expert rater (H.Adams) and the output of the method, the automated EPVS scores.
ICC is the metric most commonly used to evaluate the reliability of visual rating methods, and has also been used in previous epidemiological studies of EPVS \citep{Adams2013}. We consider it as the standard metric in our experiments.

\subsection{Saliency Maps}
\label{sec:saliency}

\begin{figure*}[t]
\centering
\includegraphics[height=8.0cm]{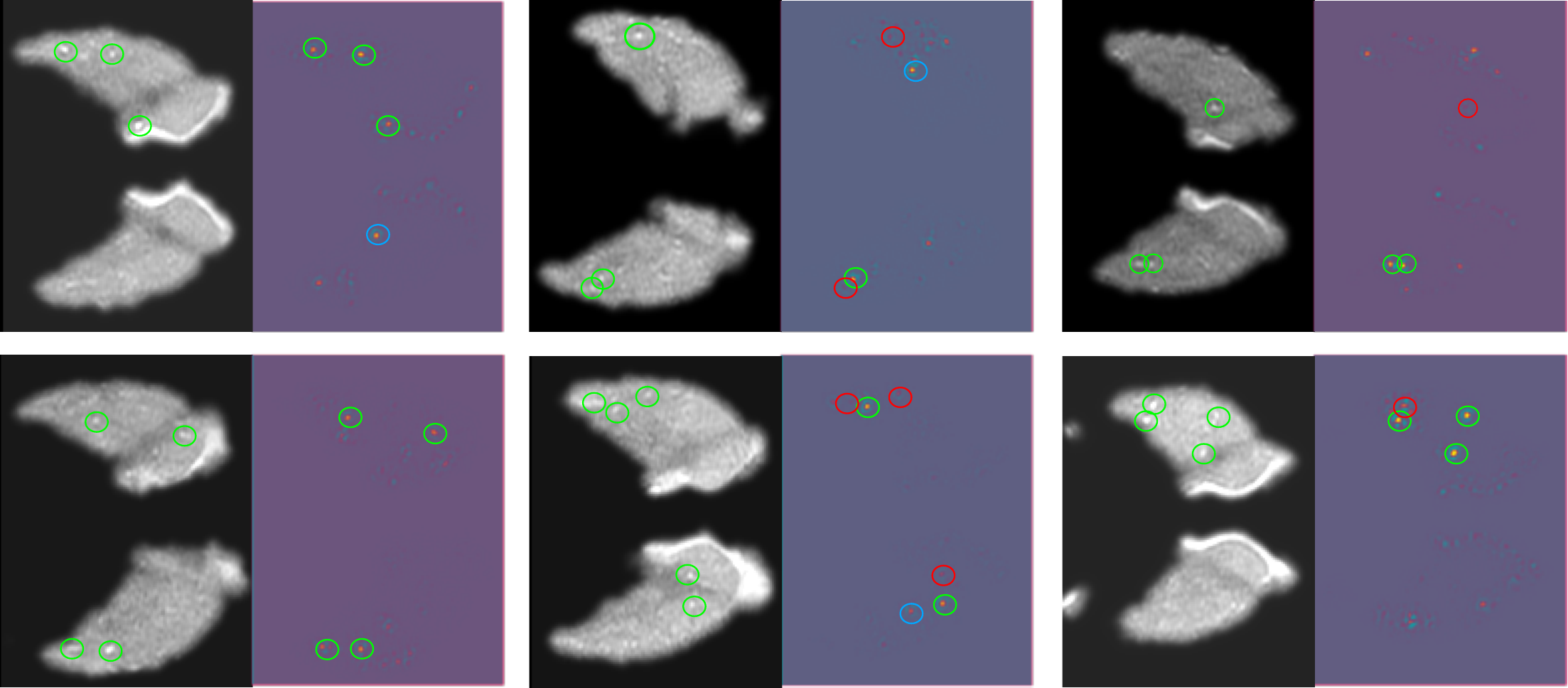}
\caption{\textbf{Examples of saliency maps.} We display the middle slices of 6 scans on the left, and the corresponding rescaled saliency maps produced by the network \citep{Simonyan2013} in right.  
On the scans, green circles highlight EPVS. On the saliency maps, regions of high activation matching with an EPVS in the scan are circle in green. When these do not match any EPVS, they are circled in blue. If a region is not activated by the presence of an EPVS, it is circled in red.}
\label{fig:saliency}
\end{figure*}

In figure \ref{fig:saliency}, we computed 6 saliency maps using our trained model (Sec. \ref{sec:regNet}). Saliency maps are computed as the derivative of the automated EPVS scores (the output of the network) with respect to the input image\citep{Simonyan2013}. Saliency maps highlight regions which contributed to the EPVS score and consequently we expect them to highlight EPVS.

After rescaling intensities of the saliency map in $[0,1]$, we circled the regions with a value higher than 0.5. Most strongly highlighted regions correspond to EPVS, although sometimes large EPVS are only slightly highlighted, while smaller-sized EPVS (that do not exceed the threshold to be counted as enlarged by the expert human rater) can be highlighted as well. In most of the cases, regions with values in $[0,0.5]$ in the saliency maps actually correspond to thin perivascular spaces.

It should be noted, however, that enlargement of perivascular space is not a 0/1 phenomenon (as a visual rating assumes) but actually happens on a continuous scale, and it is very likely that the CNN counts the EPVS in a volumetric manner. Many smaller-sized EPVS would thus not be counted by the expert human rater as ‘enlarged’ but could still slightly contribute to the total EPVS burden computed by the algorithm, hence the slightly highlighted (values in $[0,0.5]$) in the saliency maps.

Note that, while the annotator considers EPVS only in a single slice, the algorithm is considering the complete 3D volume. The number of EPVS in the annotated slice and in the total volume of the basal ganglia are strongly correlated \citep{Adams2013}. The algorithm most probably uses this correlation and locates EPVS in the total volume and scales down its output to make it match the number of EPVS in the annotated slice. We observe the same behavior in Sec. \ref{sec:occlusion}.

\subsection{Occlusion of EPVS}
\label{sec:occlusion}

\begin{figure*}[t]
\centering
\includegraphics[height=5.05cm]{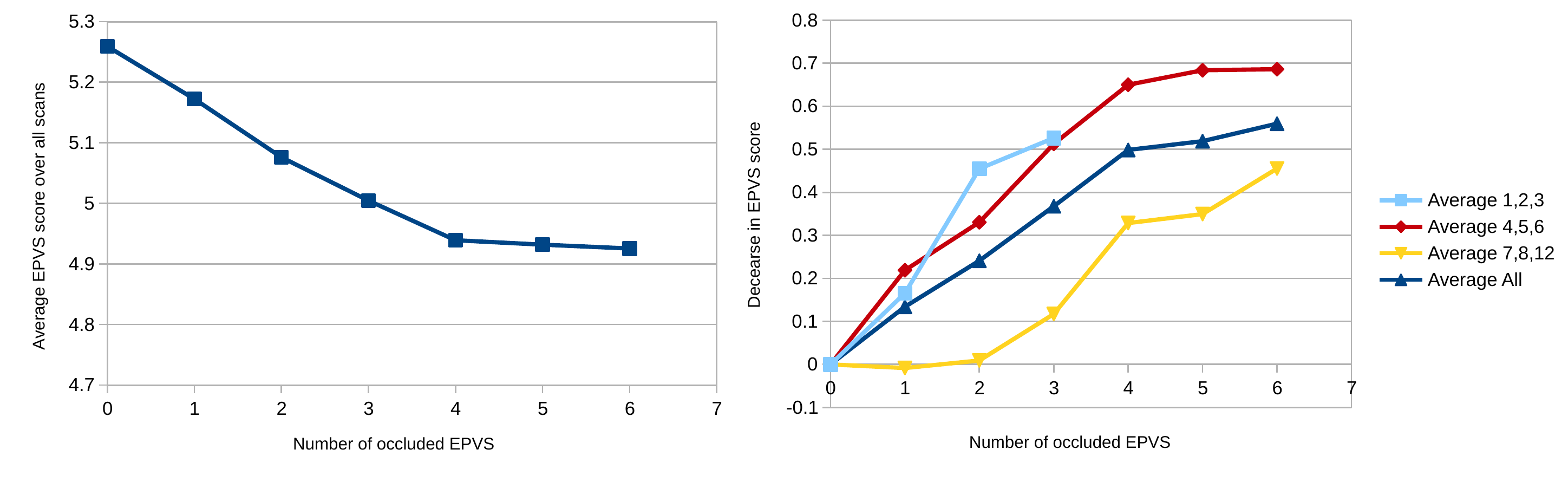}
\caption{\textbf{Predicted scores after EPVS occlusion for increasing number occluded EPVS.} On the right plot, the scores are averaged among group of scans having similar initial numbers of EPVS. For instance the light blue label stands for scans having either 1, 2 or 3 marked EPVS in the slice showing the anterior commissure. Once a scan has no EPVS to remove in the annotated slice, the predicted score stays the same. For the light blue curve, as no scans has more than 3 EPVS to occlude, the curve would stay constant after 3 EPVS. We do not plot these points. }
\label{fig:occludeRes}
\end{figure*}

In this section, we perform another experiment to verify that the algorithm learns EPVS. We use a set of 25 scans in which EPVS have been marked with a dot in the slice showing the anterior commisure (Sec. \ref{sec:data}).

The experiment consists of occluding marked EPVS with small 3D blocks (1.5x1.5x4.8 mm) of the mean intensity of the basal ganglia. We successively occlude $n$ EPVS, with $n \in [1;6]$, in all images and recompute, for each $n$, the predicted EPVS score for each image. We expect the scores to decrease as we occlude more EPVS. 

Fig. \ref{fig:occludeRes} shows the results. In the left plot, the automated scores linearly decrease as more EPVS are occluded, until four EPVS have been removed. Note that in the right plot, it seems that the automated score of scans with a lower amount of EPVS decreases quicker than for scans with many EPVS. In scans with many EPVS, the EPVS selected for occlusion may more frequently be a slightly enlarged EPVS, considered as a limit case by the algorithm and hence having a small impact on the automated score.
In the left plot, after four EPVS have been removed, the slope of the curve decreases. At that point, most of EPVS have been removed from the images, only remains images with many EPVS. 

One could expect the scores to decrease by $n$ as we occlude $n$ EPVS. The scores decrease instead by a smaller amount. The automated EPVS scores are indeed computed across the volume and scaled down to match the visual scores that were based on a single slice. Removing a single EPVS slightly affects the automated EPVS score.

\begin{figure*}
\centering
\includegraphics[height=12cm]{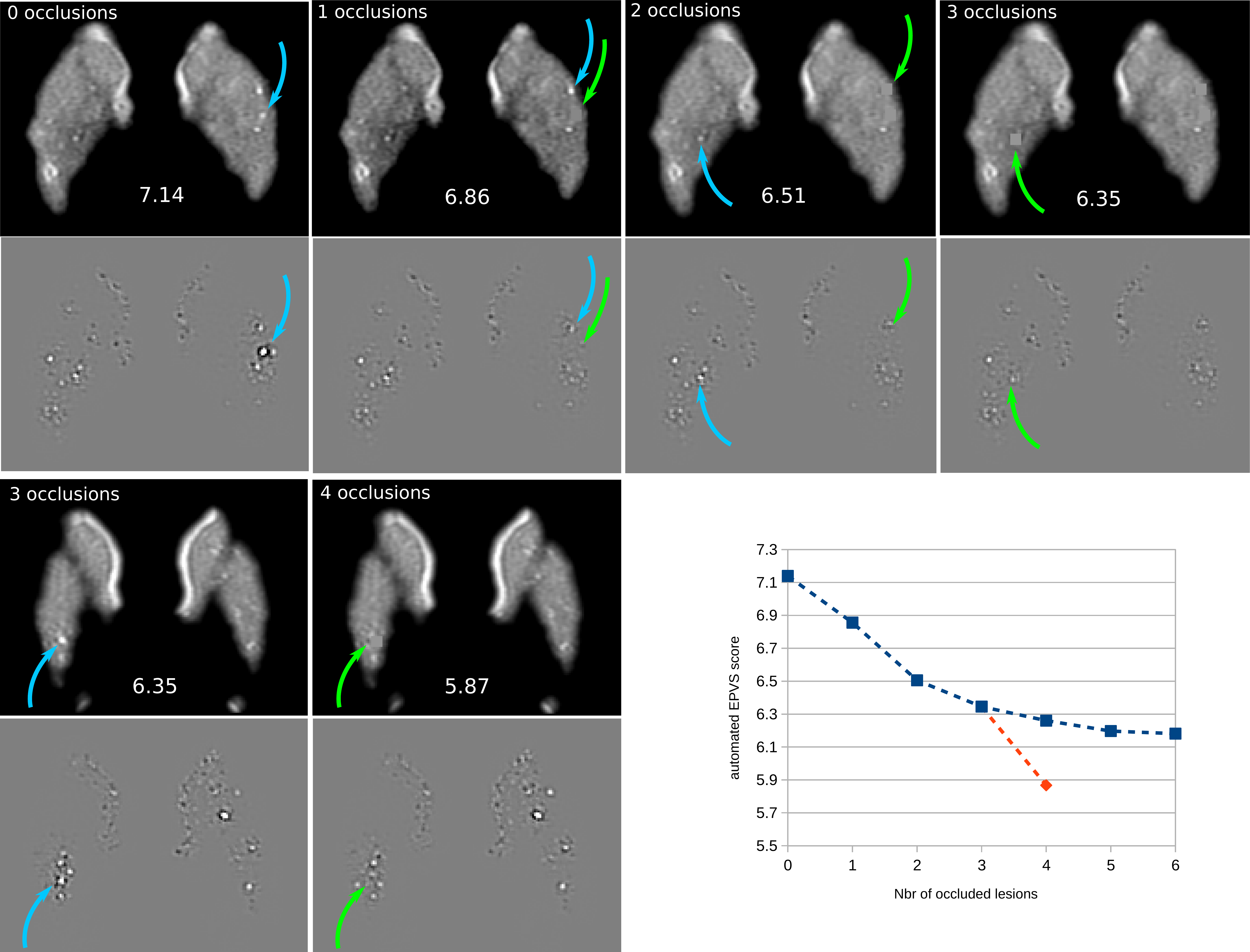}
\caption{\textbf{Occlusion in a single image.} Several lesions are progressively occluded in the same image. In the first row, we occlude lesions in the slice annotated by the expert rater. In the second row, we occlude an additional lesion in an upper slice of the same 3D image. 
The top image is the input image, and the bottom one is the corresponding saliency map (see Section \ref{sec:saliency}).
We indicated the number of occluded lesions at the top of each image, and the updated automated EPVS score in the middle. Blue arrows indicate lesions which will be occluded next. Green arrows indicate the location of lesions that have just been occluded. 
In the bottom-left of the figure, we also plot the evolution of the automated EPVS score while removing lesions. Blue is removing lesions in the annotated slice. Orange is removing the lesion in the upper slice (second row of images). Results are interpreted in Section \ref{sec:occlusion}.}

\label{fig:occlusionSaliency}
\end{figure*}

In Figure \ref{fig:occlusionSaliency}, we performed additional experiments to verify this hypothesis. As expected, we notice that occluding a lesion in the input image reduces the intensity at that location in the saliency map. However we also notice, that the more lesions are occluded in a single slice, the lower the influence on the saliency map is, and the less the automated EPVS score decreases. After removing the most obvious lesions, we actually start to occlude only slightly enlarged ones, that have a lower impact on the quantification. If we now occlude more enlarged lesions in other slices, the saliency map and automated EPVS scores are again more impacted. This confirms the hypothesis that the algorithm considers EPVS across the volume of the basal ganglia.

For comparison, we also occluded the image of Figure \ref{fig:occlusionSaliency} at random locations. We occluded 1-5 random locations in the basal ganglia, and repeated the experiment 100 times. With no occlusion, the EPVS score was 7.14. One random occlusion led to an EPVS score of 7.12 +/- 0.1 (standard deviation). This decrease is negligible in comparison to the change in EPVS score after occluding one EPVS: 6.86. Occluding five random locations led to an EPVS score of 7.10 +/- 0.28. Thus, occluding EPVS has a significant impact on the PVS score in contrast to occluding random locations. We can therefore conclude that the algorithm focuses on EPVS.

\subsection{Comparison to visual scores and to other automated approaches}
\label{sec:base}

\begin{figure*}
\centering
\includegraphics[height=6.5cm]{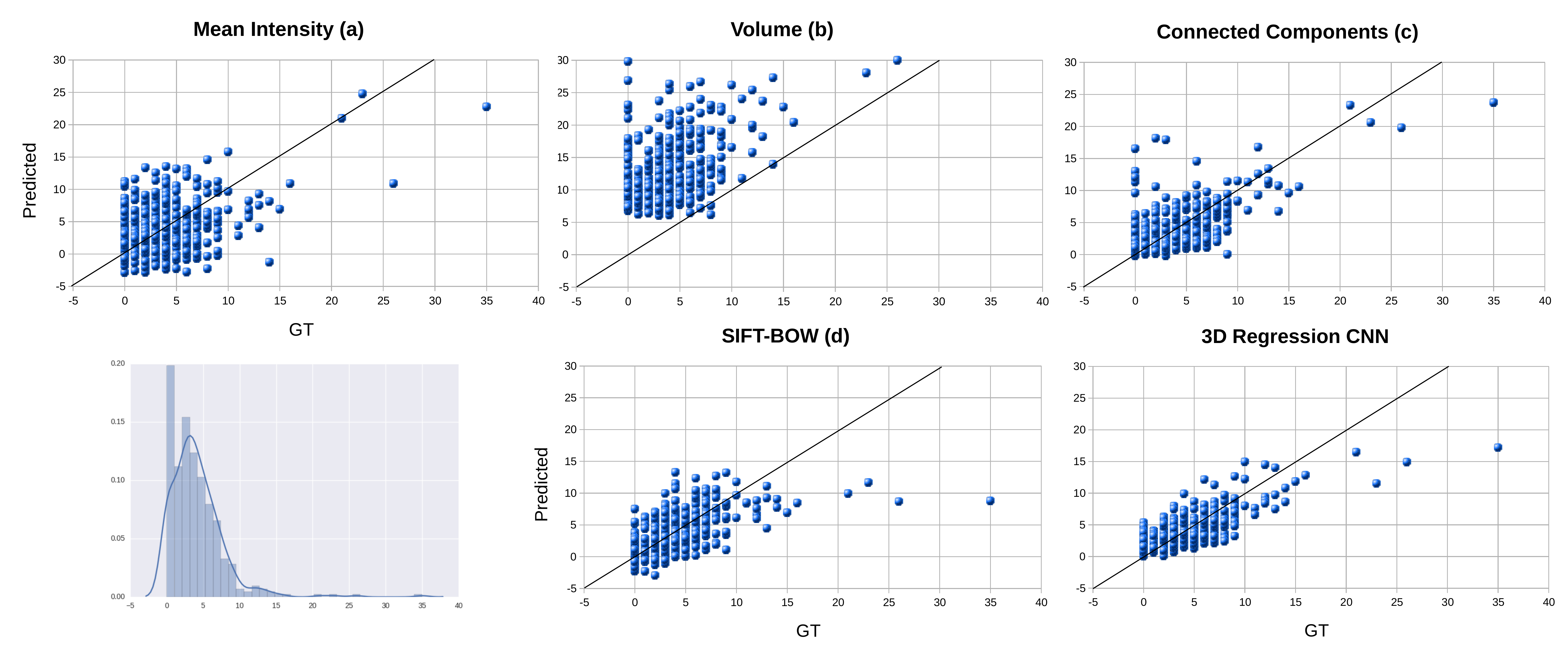}
\caption{\textbf{Regression results on the test set.} The different methods are detailed in Sec. \ref{sec:base}. The ground truths are represented on the x-axis. The predicted outputs of the methods are on the y-axis. See Table \ref{table:resultsBaselines} for correlation coefficients. On the bottom-left, we plot a histogram of the distribution of EPVS visual scores across scans.}
\label{fig:results}
\end{figure*}
\begin{table*}[h]
\setlength{\tabcolsep}{10pt}
\caption{\textbf{Correlation with expert's visual scores for the proposed method and four other more conventional approaches.} We also report the mean square error (MSE). Best performance in each column is indicated in bold.}
\begin{center}
\begin{tabular}{c c c c c c} 
\hline\rule{0pt}{12pt}
Method & Pearson & Spearman & ICC & MSE\\
\hline\rule{0pt}{12pt}
Intensity (a)	&	0.38	&	0.19	&	0.37 &  18.36 &\\
Volume (b)	&	0.47	&	0.34	&  -0.27 & 116.2 &\\
Components (c)	  &	0.63	&	0.48	&	0.63 & 9.88 &\\
SIFT-BOW (d)   &   0.57	&	0.59	&	0.55 & 10.05 &\\
3D Regression CNN	&	\textbf{0.75}	&	\textbf{0.61} &	 \textbf{0.74} & \textbf{6.14} \\
\hline\rule{0pt}{12pt}
\end{tabular}
\end{center}
\label{table:resultsBaselines}
\end{table*}

\begin{figure*}
\centering
\includegraphics[height=7cm]{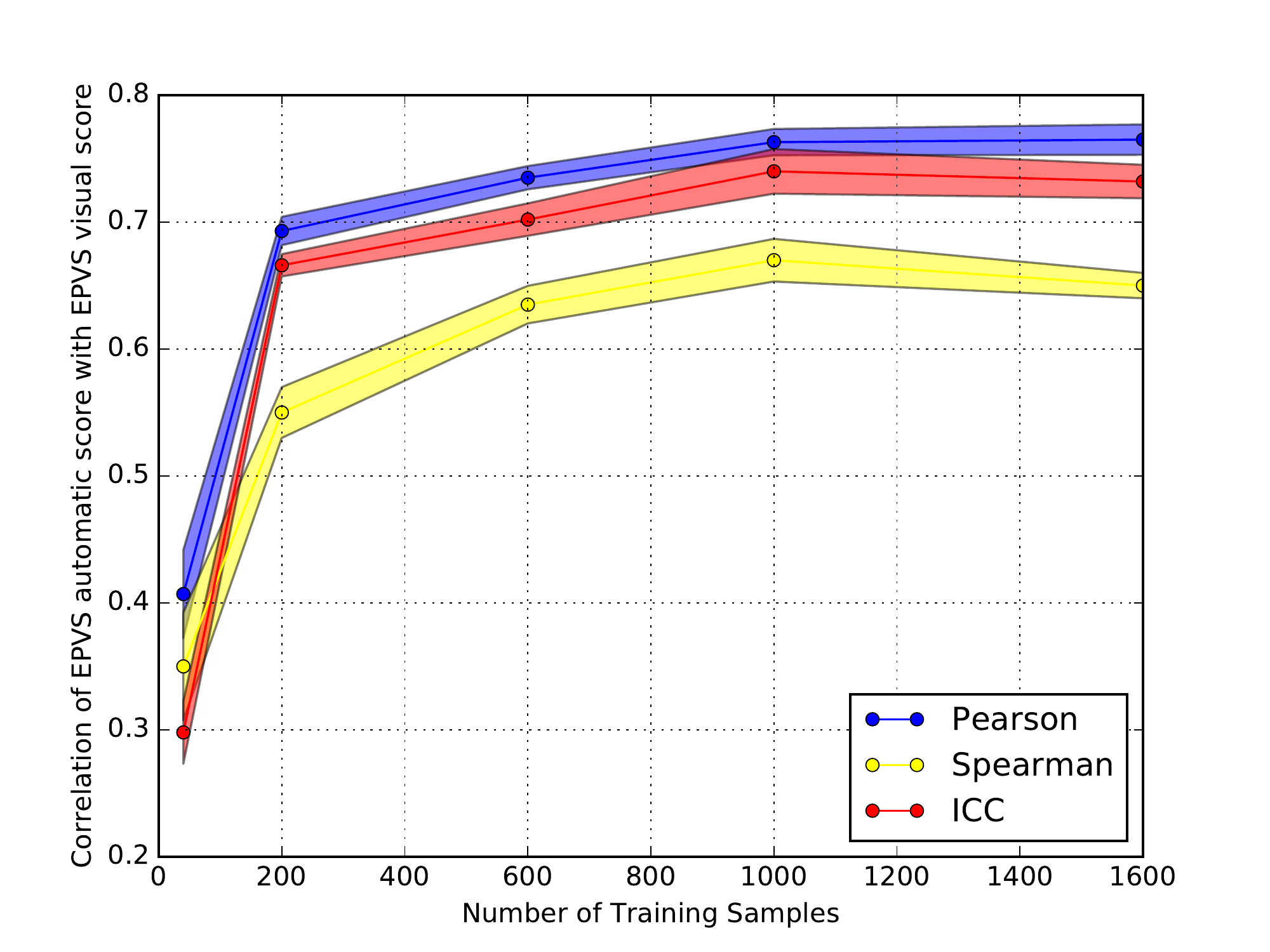}
\caption{\ textbf{Learning Curve.} The number of scans for training (80\% training set and 20\% validation set) is represented on the x-axis. Three different correlation coefficients (Pearson, Spearman, Intraclass) with visual scores are represented on the y-axis. For a given number of training samples, we average the results over 5 experiments. For each experiment, the data is randomly split into non-overlapping train, validation and test sets. Across experiments, the sets overlap (Monte Carlo cross-validation). For each point, we plot the 95\% confidence interval related to the corresponding 5 experiments.}
\label{fig:learningCurve}
\end{figure*}

\begin{table*}[h]
\setlength{\tabcolsep}{8pt}
\caption{\textbf{Intraclass Correlation Coefficent for Interrater Reliability.} A stands for the rater H. Adams, B1 is the first rating of the rater F. Dubost, and B2 the second rating of rater F. Dubost. See end Section \ref{sec:base} for more details.}
\begin{center}
\begin{tabular}{c c c c} 
\hline\rule{0pt}{12pt}
& A& B1& B2 \\
\hline\rule{0pt}{12pt}
B1& 0.70& &\\
B2& 0.68& 0.80&\\
Proposed Method& 0.80& 0.62& 0.70\\
\hline\rule{0pt}{12pt}
\end{tabular}
\end{center}
\label{table:Interrater}
\end{table*}

In this section we compare the automated scores to visual scores and demonstrate the effectiveness of our method in comparison to four other automated approaches. 

For the first series of experiments, the dataset is randomly split into the following subsets: $1289$ scans for training, $323$ for validation and $405$ for testing. The first three methods (a,b and c) quantify hyperintense regions in the MRI scans. The last method (d) is a machine learning approach similar to a state-of-the-art technique for EPVS quantification in the basal ganglia \citep{Gonzalez2016,Gonzalez2017}. These four baseline methods are particularly interesting as they cover a wide range of complexity.

The output of method (a) is simply the average of all voxels intensity values inside the ROI $S$.
Both the second (b) and third (c) method first thresholds $S$ to keep only high intensities. This threshold is optimized on the training set, without applying the intensity standardization described in Section \ref{sec:param}. We denote $S_{t}$ the thresholded image $S$. The output of (b) is the volume - the count of non-zero values - of the threshold image $S_{t}$. The output of (c) is the number of connected components in $S_{t}$.
The method (d) computes bag of visual words (BoW) features using SIFT \citep{Lowe2004} as descriptors and uses a regression forest. SIFT parameters are tuned - by visual assessment - to highlight EPVS on the training set. 2D SIFT are computed in each of the 15 slices surrounding the slice annotated by clinicians. In our experiments, using more surrounding slices proved to be too complex for the model, which would then fail to learn the aimed correlation. The number of words in the BoW dictionary was set to 100 for each slice. Concatenating the feature vectors of each slice yielded better results than averaging these vectors. The BoW features for the entire volume are therefore vectors of $15*100=1500$ elements. The regression forest has 3000 trees and a maximum depth of 50 nodes.

For all these other automated approaches, the regression results need to be rescaled to be able to compute the ICC. We apply a linear transformation to the outputs. The predicted values can consequently become negative. The parameters of this transformation are optimized to maximize the ICC on the validation set.

We report the results of this experiment in Table \ref{table:resultsBaselines}. The regression network performs best for all measures and outperforms the other methods by a large margin - more than 0.10 - for both Pearson correlation and ICC. Our method performs significantly better than all four baselines (William's test, p-value $<$ 0.00001 for baselines (a), (b), (d) and $<$ 0.01 for baseline (c)). Methods (c) and (d) are the strongest baselines

Fig.\ref{fig:results} presents scatter plots of the estimated outputs for each method. We notice that method (c) sometimes strongly overestimates the number of EPVS in scans with no EPVS. Such errors do not happen with our regression network. On the other hand, method (d), and to a lesser extent the proposed method, have a tendency to underestimate EPVS in scans with the largest amounts of EPVS. A possible explanation for this underestimation is that in case of a larger number of EPVS, the chance of having lesions close to each other is higher. This makes the detection more challenging. Several very close EPVS may appear similar to a single larger EPVS in other scans. 

Note that despite its simplicity, method (c) performs reasonably well, especially in comparison with the random forest (d), which is much more complex (more parameters). However, note that the performance metrics of method (c) as displayed in Table \ref{table:resultsBaselines} are strongly influenced by few scans having many EPVS (see Figure \ref{fig:results}). If we ignore these scans and recompute the ICC for scans with only 20 EPVS or less, method (c) drops to 0.48 ICC and 11.02 MSE while method (d) gets to 0.59 ICC and 9.22 MSE and the proposed method is at 0.68 ICC and 6.74 MSE.

In the experiments described above, we have demonstrated that the scores predicted by our algorithm have a good to excellent (according to \cite{Cicchetti1994} guidelines) correlation with the scores of a single expert rater (H. Adams). However, as the algorithm is trained with the scores of this same rater, its predictions may be biased. 

To verify this, we evaluated the performance of our algorithm on a smaller set annotated by two raters (H. Adams and F. Dubost) (see Section \ref{sec:data}). For this experiment we trained the algorithm on a training set (training + validation) of 1600 scans and a test set of 400 scans. Table \ref{table:Interrater} shows the results.

\subsection{Learning Curve}
\label{sec:learningCurve}

In this section we study how the number of annotated scans used for optimization influences the performance of our automated quantification method.

We train our network using different subsets of the 2017 MRI scans described in Sec. \ref{sec:method}. We perform experiments using 5 different sizes for the training set. For a fixed number of training scans, we repeat the experiment $5$ times with different randomly drawn train/test splits of the data. This results in $5*5=25$ experiments with different random train/test splits of the data. Fig. \ref{fig:learningCurve} shows the results of the experiment. In the training set size, we count both training (80\%) and validation (20\%) sets.

Even with a relatively small training set size (200 scans) our method performs well: the correlation between the automated and visual scores reaches an ICC of 0.66. Our model reaches its best performance (ICC of $0.74 \Mypm 0.044$) with 1000 training scans. Using more scans does not bring further improvement. Using only a few training scans (40) leads to a significant drop in performance (ICC of 0.30) with higher standard deviation.

\subsection{Analysis of Network Parameters}
\label{sec:NetParam}

\begin{table*}[h]
\setlength{\tabcolsep}{8pt}
\caption{\textbf{Network parameters and Corresponding Results.} See Sec. \ref{sec:NetParam} for details. $\star$ indicates the proposed method. In $\star\star$, the network has only half of the features of the other variants in this table. The best results per category of experiment are in bold. }
\begin{center}
\begin{tabular}{c c c c c c c c c c c c c} 
\hline\rule{0pt}{12pt}
MNI& Feat1stL & FlipX & FlipY & FlipZ & FC & Loss & Blocks & Conv/Block & ICC & MSE\\
\hline\rule{0pt}{12pt}
$\star$ 1&32&1&1&1&2*2000&MSE&2&4&0.783&4.37\\
\hline\rule{0pt}{12pt}
0&32&1&1&1&2*2000&MSE&2&4&0.771&4.99\\
\hline\rule{0pt}{12pt}
1&32&1&1&1&2*2000&MCE&2&4&\textbf{0.751}&6.11\\
1&32&1&1&1&2*2000&MFE&2&4&0.708&\textbf{5.76}\\
1&32&1&1&1&2*2000&tukey&2&4&did not&converge&\\
1&32&1&1&1&2*2000&RMSE&2&4&did not&converge&\\
\hline\rule{0pt}{12pt}
1&32&1&1&1&2*2000&MSE&1&4&0.807&4.76\\
1&32&1&1&1&2*2000&MSE&2&3&0.805&4.93\\
1&32&1&1&1&2*2000&MSE&2&2&0.808&5.03\\
1&32&1&1&1&2*2000&MSE&2&1&0.803&4.85\\
$\star\star$1&16&1&1&1&2*2000&MSE&3&4&0.767&5.64\\
\hdashline[0.5pt/3pt]\rule{0pt}{12pt}
1&16&1&1&1&2*2000&MSE&2&1&0.776&5.17\\
1&32&1&1&1&2*2000&MSE&2&1&\textbf{0.803}&\textbf{4.85}\\
1&64&1&1&1&2*2000&MSE&2&1&0.780&5.14\\
\hdashline[0.5pt/3pt]\rule{0pt}{12pt}
1&32&1&1&1&1*2000&MSE&2&4&0.781&5.06\\
1&32&1&1&1&0&MSE&2&4&0.788&4.76\\
\hline\rule{0pt}{12pt}
1&32&0&1&0&2*2000&MSE&2&4&\textbf{0.787}&\textbf{4.65}\\
1&32&0&0&0&2*2000&MSE&2&4&0.742&5.88\\
1&32&No&Data&Augm&2*2000&MSE&2&4&0.742&6.23\\
\hline\rule{0pt}{12pt}
\end{tabular}
\end{center}
\label{table:NetParam}
\end{table*}

\begin{figure*}[t]
\centering
\includegraphics[height=7.05cm]{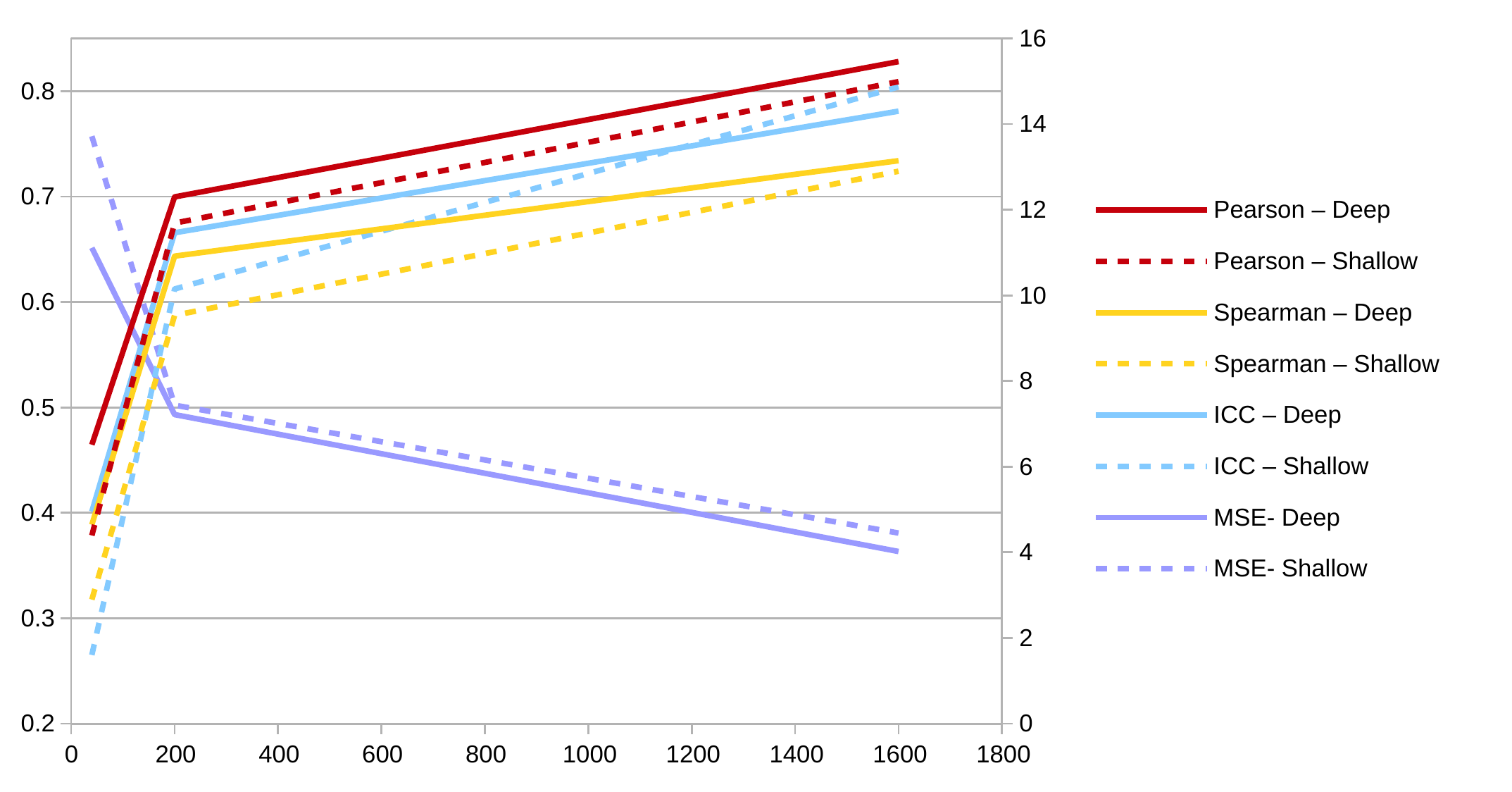}
\caption{\textbf{Learning Curve of shallow and deep network.}The number of training and validation scans is displayed on the x-axis. The correlations coefficients (Pearson, Spearman and ICC) between automated and visual scores are displayed on the left y-axis (the scale ranges from 0.2 to 0.85). The MSE between automated and visual scores is displayed on the right y-axis. Solid lines are used for the deep network, and dotted lines for the shallow network.}
\label{fig:shallowlearningCurve}
\end{figure*}

In this section, we investigate the influence of several parameters of the model. Table \ref{table:NetParam} summarizes a set of experiments performed on the same split of training, validation and testing set, which sizes are 1289 scans, 323 and 405 respectively. In this series of experiments the varying parameters are: registration to MNI space (MNI); number of features in the first layer (Feat1stL); for the data augmentation, flipping scans in the direction of the sagittal axis (FlipX), the left-right axis (FlipY), the longitudinal axis (FlipZ); the layout of the fully connected layer (FC), where e.g. 2*2000 means 2 layers of 2000 neurons each; the loss (Loss), where MSE stands for mean square error, MCE for mean cubic error, MQE for mean quartic error, Tukey for Tukey's biweight and RSME for root mean square error. Blocks is the number of convolutional blocks as described in Sec. \ref{sec:method} and Conv/Block is the number of convolutional layers per block. ICC and MSE are the metrics we computed on the test set. Note that we conducted these experiments a posteriori and did not use these results to tune the parameters of the method for the experiments in sections \ref{sec:base}, \ref{sec:learningCurve}, \ref{sec:repro} and \ref{sec:ageCor}.

Table \ref{table:NetParam} is separated in several categories of experiments. The first line shows the algorithm implemented in this article. On the second line we notice that registering to MNI spaces does not provide a large improvement. In the third category, we investigate several loss functions. MSE provides a better performance. In the fourth category we investigate different architectures. Reducing the number of convolutional layers or fully connected layers does not bring a large difference, neither does changing the number of features in the first layer. To perform the experiment with three blocks, we halved the number of features maps in each layer. This architecture yields worse results than shallower architectures. The last category investigates different levels of data augmentation. The most important augmentation is flipping the images in the y-axis, which is an anatomically plausible augmentation. Other forms of data augmentation bring no improvement in this scenario and can make the training process more difficult and slower.

Overall, in this problem setting, registering to MNI is not necessary, MSE is the loss of choice, architecture changes do not bring significant differences but one could prefer using a smaller network for faster training, and the best augmentation is flipping in anatomically plausible directions.       

We noticed in table \ref{table:NetParam} that, considering the ICC, shallower networks perform similar to deeper ones in this problem (in regards to the MSE, the proposed deep network performs slighted better though). We investigate the behavior of these shallower models for smaller amount of training samples. Figure \ref{fig:shallowlearningCurve} shows a comparison of the learning curves of a deep network (as implemented in this article) and a shallow network with two blocks and a single convolutional layer per block (see table \ref{table:NetParam} and Sec. \ref{sec:method}). The deep network performs slightly better and the difference in performance is larger for smaller training sets.

\begin{figure*}
\centering
\includegraphics[height=7cm]{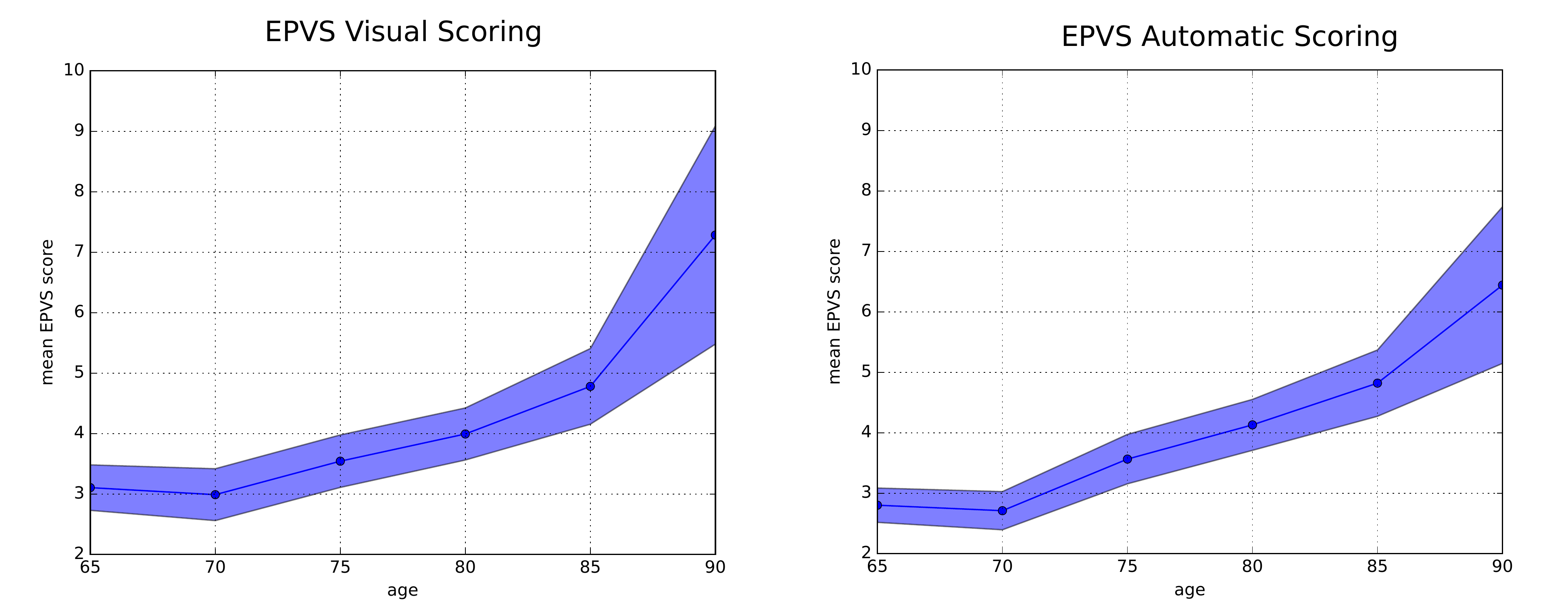}
\caption{\textbf{EPVS scores as a function of age.} We show the mean EPVS scores and 95\% confidence interval per 5 years, for visual (left) and automated (right).}
\label{fig:corrAge}
\end{figure*}

\subsection{Reproducibility}
\label{sec:repro}

In order to evaluate the reproducibility of our automated EPVS scoring method, we run our algorithm on the reproducibility set described in Sec. \ref{sec:data}. In this experiment we consider two versions of our model. For each version, we trained a set of 5 networks with randomly selected training sets of scans. For both versions, we actually use the same networks as in the learning curve experiments (Fig. \ref{fig:learningCurve}). In the first version, the networks have been optimized using 1000 scans and yields a ICC of 0.740 $\Mypm$ 0.044 with visual scores from the human rater. In the second version, the networks have been optimized only with 40 scans and yields an ICC of 0.298 $\Mypm$ 0.062 with the visual scores. On the reproducibility set, the first model yields an ICC of 0.93 $\Mypm$ 0.02 between the first and second sets of scans. The second model yields an ICC of 0.83 $\Mypm$ 0.011. According to \cite{Cicchetti1994} guidelines , both models have an excellent correlation. \cite{Adams2013} reported an intrarater agreement of 0.80 ICC for EPVS visual scoring in the basal ganglia. In our study, the second rater also had an intrarater agreement of 0.80 ICC (Section \ref{sec:base}). From this comparison we can conclude that our automated EPVS scoring appears to be more reproducible than visual scoring.

\subsection{Correlation with Age}
\label{sec:ageCor}

Now that we have demonstrated the performance of our approach in comparison with other automated approaches and human visual scores, we investigate the correlation of our automated EPVS scores with clinical factors. EPVS have been shown to correlate with age \citep{Potter2015a}. We consider correlations between age and visual EPVS scores from human raters (a), and between age and automated EPVS scores (b). We split our dataset into a training set of 1000 scans and a testing set of the remaining 1000 scans. We use the training set to optimize the parameters of our automated scoring algorithm. For (a) and (b), we perform a zero-inflated negative binomial regression. The model is zero-inflated to take into account the over-representation of participants with no EPVS (see EPVS distribution across participants in Fig. \ref{fig:results}). The per-decade odds ratio and 95\% confidence interval are for (a) 1.30 $\Mypm$ 0.08 and for (b) 1.34 $\Mypm$ 0.07. Fig. \ref{fig:corrAge} shows the trends of increasing EPVS scores with age, which are very similar for automated and visual scores.

%------------------------------------------------

\section{Discussion}

We showed that our regression network indeed focuses on EPVS to compute the automated scores, although no information about the location of these lesions had been given during training. This automated scoring has a good agreement with the visual scoring performed by a single expert rater, is highly reproducible, and significantly outperforms the scoring of the four more conventional methods we compared to. 

Few other papers addressed EPVS quantification. In contrast with our approach, \cite{Gonzalez2017} formulated the problem as a binary classification where a threshold is set to $t=10$ EPVS to differentiate between the severe or mild presence of EPVS. The authors use bag of visual words and SIFT features \citep{Lowe2004}, similar to our baseline method (d), and achieve an accuracy of $82\%$ on a test set of $80$ scans. The regression approach as presented in our paper provides a much finer - and therefore likely more relevant - quantification than this binary classification. In addition, in our experiments, the regression network yields much better results than the bag of words with SIFT approach (Table \ref{table:resultsBaselines}). In Figure \ref{fig:results}, the bag of word approach (d) is also more spread along the second principal component, meaning that this method is in average less precise in its quantification (high mean square error). This matches with the mean square errors reported in Table \ref{table:resultsBaselines}.

More recently, the same authors \citep{Ballerini2016} used methods based on vessel enhancement filtering, and reported a Spearman correlation of $0.75$ with a 5-category EPVS ranking (the Potter scale, \cite{Potter2015b}) in the centrum semiovale. Our method achieves a Pearson correlation of $0.763 \Mypm 0.026$ and a Spearman correlation of $0.670 \Mypm 0.042$ with visual scoring in the basal ganglia. These results cannot directly be compared as the regions, visual scoring systems, and datasets are different.
A possible advantage of the visual EPVS score used in our work \citep{Adams2013} with respect to the Potter scale \citep{Potter2015b}, is that it provides  a finer quantification. In our study population, the majority of images would fall into the first 2 categories of the Potter scale  (0 EPVS and 1-10 EPVS), while the score of \cite{Adams2013} allows further separation.

\cite{Ramirez2015} developed interactive segmentation methods based on intensity thresholding. The authors show good results but need the intervention of a human rater, which in large datasets is an important drawback. Our method is fully automated.
\cite{Park2016} proposed an automated EPVS segmentation method based on Haar-like features. This method reaches up to 64\% Dice coefficient with ground truth annotations. This approach was exclusively evaluated on 7 Tesla MRI scans, needs a large amount of pixel-wise annotations for training, and was only evaluated on a dataset of 17 young healthy subjects. We evaluated our method on the Rotterdam Scan Study \citep{Ikram2015}, a population-based study in middle aged and elderly subjects. The elderly subjects are more prone to cerebral small vessel diseases, and may have other types of brain lesions, similar to EPVS (e.g. lacunar infarcts). This makes the exclusive quantification of EPVS more challenging on our dataset, but also closer to the clinical need.

Several other learning-based approaches to counting objects in images have been proposed in the literature, mostly in case of 2D images. These techniques also often need labels about the location of the target objects. 
\cite{Lempitsky2010} proposed a supervised learning method to count objects in images. However their method is based on density map regression and relies on dot annotations for training.
More recently, \cite{Walach2016} proposed a convolutional neural network with boosting and selective sampling for cell and pedestrian counting. Their method is also base on density map regression and needs dot annotations.
\cite{Ren2016} proposed a method to jointly count and segment instances in 2D images. They combined a recurrent neural network with an attention model. However the method needs a pixel-wise ground truth for its segmentation component.
\cite{Segui2015} proposed a convolution neural network for counting handwritten digits and pedestrians. The network are optimized for classification with weak global labels: the number of instances of the target object. This work is closer to our method, as we also use weak global labels. However, we use regression networks.
All these method were evaluated only on 2D tasks. For instance, overcoming occlusions is one of the main difficulties tackled in pedestrian counting, a problem which does not occur in case of 3D volumes. 

Our method is both reproducible (0.93 ICC) and agrees well with the visual scores of the expert human rater it has been trained on: the correlation between the automated and visual scores is 0.74 ICC, which is in between interrater agreement (0.62 in \cite{Adams2013}, and 0.68 and 0.70 in our study (Table \ref{table:Interrater})) and intrarater agreement (0.80 in both in \cite{Adams2013} and our study (Table \ref{table:Interrater})). Furthermore, the correlation between the automated scores and the visual scores of a second expert human rater - which have not been seen during training - is similar to that of the interrater agreeement (Table \ref{table:Interrater}). Therefore, we believe our method is sufficiently precise and robust to perform automated EPVS quantification in large scale clinical research. The processing time stays low enough: $440$ ms on GPU per scan given to the regression network. However, as all images in our database were acquired with a single scanner, for application in different data it would need to be evaluated on a multi-center dataset to further verify its robustness.
Additionally, our method was exclusively evaluated in the basal ganglia, as perivascular spaces in this region are suggested to be most clinically relevant \citep{Potter2015b}. In other EPVS research studies \citep{Adams2015,Ikram2015,Maillard2016,Hilal2013}, EPVS can also be visually scored in other brain regions such as centrum semiovale, hippocampus and midbrain \citep{Adams2013}. This is particularly relevant as the location of EPVS is thought to differ with etiology and even relate to different clinical outcomes \citep{Banerjee2017,Charidimou2017}. We expect our method to perform similarly in other brain regions. 

Contrary to EPVS visual scoring, we quantify the EPVS in the entire ROI volume and not only in a single slice. However it has been shown \citep{Adams2015} that the visual EPVS score in a slice of the basal ganglia is highly correlated to the EPVS visual score in the entire volume. The results from experiments with occlusion suggest that our method uses this correlation by detecting EPVS in the whole volume and scaling the score down to match the visual scores done in a single slice. The automated scores are more robust than visual ones in this regard. Training a classifier on visual scores of the whole basal ganglia volume could provide an even more robust approach and could prove itself useful to investigate more subtle correlations with clinical factors.

In this work, we did not limit our input to the visually scored slice. The human rater indeed uses information from more than just one slice to discriminate EPVS from similarly appearing brain lesions, and we expect the network to benefit from this information as well. Besides, we expect that quantifying EPVS in the entire basal ganglia, fusing information from multiple slices, is more reliable than only quantifying them in a single slice.

In Table \ref{table:Interrater}, while the correlation between the automated scores and the visual scores of the second rater (F. Dubost) is slightly lower than the correlation between both raters (F. Dubost and H.Adams), it is still higher than the interrater ICC reported in \cite{Adams2013}. Overall, we believe that this table shows that we automatized the first rater (H. Adams), with interrater and intrarater reliabilities similar to that of expert human raters.

Looking at the learning curve (Fig \ref{fig:learningCurve}), it seems that the performance of the network does not improve when training on more than 1000 images. This could mean that either this is the maximum achievable performance using this ground truth or that increasing the complexity of the network (by adding layers and feature maps) could still lead to an increase in performance. However the experiments conducted in section \ref{sec:param} suggest that a similar performance can be achieve by shallower networks. Though, shallower networks seem to perform worse for small training sets. More regularization (Dropout, L1 or L2) may help to reduce the drop in performance (for both deep and shallow networks) when training on small amount of samples.

In theory, we think that the performance of the network could be further boosted with e.g. attention mechanisms \citep{Mnih2014}, given highly accurate ground truth labels. However, we cannot expect any methods trained on ratings of a single rater to perform better than intra-rater agreement (here ICC of 0.8). In several cases (see Table \ref{table:NetParam}) our prediction reaches this level of agreement with the expert’s scores. That is why we did not experiment with more complicated methods: with the current ground truth based on visual assessment, we can not expect nor would we be able to meaningfully evaluate any further performance gain.

The large size of the required training set could be seen as an obstacle to the clinical application of the automated scoring method. However, although our best performance is achieved with a training set of 1000 scans, training with 200 scans already provides a good performance. We believe this method can be extended to and would be useful for other large clinical and population-based studies such as ADNI \citep{Jack2008}, UK Biobank \citep{Sudlow2015} and German National Cohort \citep{Ahrens2014}.

\section{Conclusion}

We presented a novel regression method to automatically quantify the amount of enlarged perivascular spaces in the  basal ganglia in brain MRI. We validated our approach on 2000 brain MRI scans (using different sizes for the testing set, up to a maximum of 1960 scans). Our method significantly outperforms four other more conventional automated approaches. The agreement with visual scoring (ICC of 0.74) is higher than the inter-observer agreements (ICC of 0.68 and 0.70). The scan-rescan reproducibility is very high (ICC of 0.93), compared to intra-observer agreement (ICC of 0.80). Our result are relatively robust across network architectures. We also demonstrated that the  automated EPVS scores correlate with age, similarly to the visual EPVS scores. We believe that this method can replace visual scoring of EPVS in epidemiological and clinical studies.

\section{Acknowledgments}
This research was funded by The Netherlands Organisation for Health Research and Development (ZonMw) Project 104003005.

%\section*{References}

\bibliography{MedIA2018_paper}

\end{document}